%% file: main.tex
\documentclass{article}

\usepackage{nhn_preprint,times}
\input{math_commands.tex}

\usepackage[utf8]{inputenc}
\usepackage[T1]{fontenc}
\usepackage{hyperref}
\usepackage{xurl}
\usepackage{booktabs}
\usepackage{makecell}
\usepackage{multirow}
\usepackage{graphicx}
\usepackage{wrapfig}
\usepackage{float}
\usepackage{amsfonts}
\usepackage{nicefrac}
\usepackage{microtype}
\usepackage{xcolor}
\usepackage{amsmath}
\usepackage{amssymb}
\usepackage{amsthm}
\usepackage{bm}
\hypersetup{
  pdftitle={Neuromotor Hierarchy Network: Physiological Inductive Biases for Robust Generalization in sEMG Decoding},
  pdfauthor={He Wang, Hongyuan Qi, Zhaoxian Zhang, Jinbin Luo, Linyi He, Mehul Motani, Changsheng Wu}
}

\title{Neuromotor Hierarchy Network: Physiological Inductive Biases for Robust Generalization in sEMG Decoding}

\author{He Wang \quad Hongyuan Qi \quad Zhaoxian Zhang \quad Jinbin Luo\\
\bfseries Linyi He \quad Mehul Motani \quad Changsheng Wu\\
\normalfont National University of Singapore}

\begin{document}
\maketitle

\begin{abstract}
Surface electromyography (sEMG) provides a wearable, noninvasive interface to
neuromuscular activity for movement decoding and
human-computer interaction. Population-scale decoding remains difficult
because the relationship between sEMG and neuromuscular activity varies
across users and sessions, while task-relevant dynamics span channels and
multiple timescales. Learning waveform-to-output mappings from task labels
leaves the distinction between recording variability and coordinated motor
activity implicit. We introduce the Neuromotor Hierarchy Network (NHN),
which learns a compact latent neuromotor state from task supervision
to represent task-relevant neuromuscular coordination.
NHN constructs this latent state
through a hierarchy inspired by neuromotor organization.
It adapts recording statistics while preserving relative intensity.
Its spatiotemporal encoder uses parameter-efficient channel interactions
and modulates features with multi-timescale history.
The resulting features yield candidate activations of learned motor primitives,
which are temporally integrated and continuously weighted to form the state. Theoretical analysis
characterizes the efficiency, temporal behavior, and optimization
of NHN's core mechanisms.
We evaluate the architecture for both continuous
hand-pose estimation on \emph{emg2pose} and touch-typing recognition on \emph{emg2qwerty}. On
\emph{emg2pose}, NHN reduces user-averaged angular error by 0.52\% to 2.84\%
across all three generalization splits in both Regression and Tracking
relative to Hadidi et al.'s best task-specific variants, using
48.42\% to 48.51\% fewer parameters.
On \emph{emg2qwerty}, NHN reduces beam-search character error rate by
19.40\% zero-shot and 30.42\% after fine-tuning relative to SplashNet-Upscale,
using 65.86\% fewer parameters.
Physiology-guided inference of a latent neuromotor state supports
parameter-efficient sEMG decoding.
\end{abstract}

\input{sections/introduction}
\input{sections/related_work}
\input{sections/method}

\input{sections/experiments_results}

\section*{AI use statement}

We used large language models extensively for iterative manuscript
revision, including language polishing and improvements to clarity
and readability, as well as for assistance with code writing and
debugging during implementation. We also used a generative image
tool to produce an initial draft of the anatomical schematic in
Figure~1, which we subsequently modified and refined manually to
better reflect the relevant physiological structures. We reviewed
the AI-assisted text for technical accuracy and reviewed and tested
the AI-assisted code before using it in our experiments. The authors
take responsibility for the final manuscript, figures, code, and
reported results, including all content developed with AI assistance.

\section*{Ethics statement}

We use the publicly available emg2pose and emg2qwerty datasets and collected
no additional human-participant data. The original studies report ethics
approval and participant informed consent
\citep{salter2024emg2pose,sivakumar2024emg2qwerty}.
We use both datasets in accordance with their respective licenses.

\section*{Reproducibility statement}

We will release the complete codebase used for training,
subject-specific fine-tuning, and the analyses reported in this work.

\nocite{*}
\bibliography{references_nhn}
\bibliographystyle{nhn_references}

\end{document}

%% file: math_commands.tex
\def\vmu{{\bm{\mu}}}
\def\vd{{\bm{d}}}
\def\vf{{\bm{f}}}
\def\vp{{\bm{p}}}
\def\vq{{\bm{q}}}
\def\vs{{\bm{s}}}
\def\vu{{\bm{u}}}
\def\vx{{\bm{x}}}
\def\vy{{\bm{y}}}
\def\mI{{\bm{I}}}
\def\mW{{\bm{W}}}
\def\mSigma{{\bm{\Sigma}}}

%% file: sections/introduction.tex
\section{Introduction}
\label{sec:intro}

Surface electromyography (sEMG) supports prosthetic control, gesture input, continuous hand-pose estimation,
and touch-typing recognition without cameras or implanted sensors
\citep{atzori2014electromyography,farina2014extraction,kaifosh2025generic}.
Population-scale datasets pair wrist sEMG with diverse hand kinematics
\citep{salter2024emg2pose} or natural typing \citep{sivakumar2024emg2qwerty}
to support cross-user learning. Decoders must resolve fine-grained activity
while limiting sensitivity to waveform variation across users and sessions.
Resource-constrained wearables also require parameter efficiency.

Larger training populations and stronger temporal models have advanced pose
and typing decoding \citep{kaifosh2025generic,mehlman2025scaling,
hadidi2026reevaluating}. Progress also comes from constraining what these models
need to learn. SplashNet adapts signal statistics and shares weights between
the two hands to improve typing efficiency
\citep{hadidi2025splashnet}, while EMBridge \citep{cui2026embridge} transfers pose structure to electromyographic
representations through cross-modal learning.
These advances leave open how to organize a compact representation of
task-relevant neuromuscular coordination while limiting sensitivity to
recording-specific variation. The physiology of sEMG generation can guide this organization.

\begin{wrapfigure}{R}{0.40\textwidth}
\centering
\includegraphics[width=0.95\linewidth]{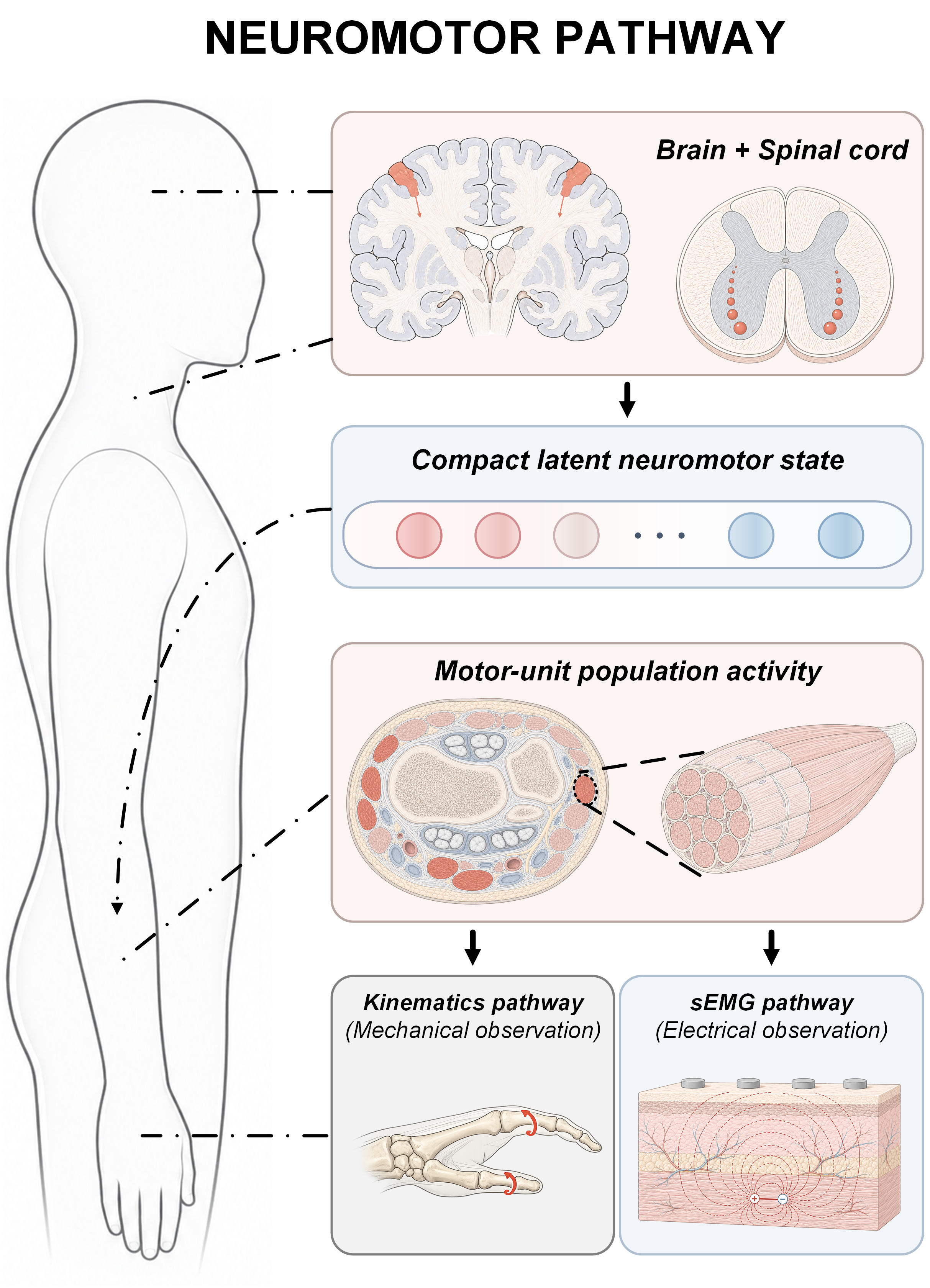}
\caption{Physiological inspiration. Neuromotor coordination links hand motion
and sEMG, motivating a compact latent neuromotor state.}
\label{fig:neuromotor_inspiration}
\end{wrapfigure}
The neuromotor pathway links motor-neuron activity to coordinated muscle
activation, giving rise to both movement and surface electrical signals
(Figure~\ref{fig:neuromotor_inspiration}). At the
neural level, motor neurons integrate synaptic input over time, while recruitment
and discharge-rate modulation shape graded motor output
\citep{negro2016common,powers2017synaptic,heckman2012motor}. Their discharges evoke
action potentials in muscle fibers, initiating contraction through
excitation-contraction coupling \citep{heckman2012motor}. Across muscles,
activation forms coordinated patterns that vary in strength and timing,
producing forces that act through the musculoskeletal system to generate
movement \citep{davella2003combinations,bizzi2013neural}. The same muscle-fiber
action potentials also give rise to the electrical signals recorded at the skin,
where tissue filtering and volume conduction shape the mixture observed by
surface electrodes. Anatomy, electrode placement, contact, and gain affect the scale and spatial distribution of the recorded waveform
\citep{farina2002influence,farina2014extraction,farina2025extraction}.

Thus, sEMG and hand kinematics share a neuromotor origin, but their relationship
varies with recording conditions, anatomy, and individual motor control
\citep{farina2014extraction,kaifosh2025generic}. Aligned latent representations
support cross-session and cross-individual decoding in cortical and
sEMG studies \citep{gallego2020longterm,safaie2023preserved,almashhadani2025toward}.
We hypothesize that a compact
latent neuromotor state can accommodate this variability by accounting for
measurement effects and representing coordination through physiologically
structured components \citep{davella2003combinations}. Their activation strength
and timing can vary within a shared representation to accommodate
individual motor patterns and reduce reliance on recording- or user-specific waveforms.

The Neuromotor Hierarchy Network (NHN) implements this inference process
(Figure~\ref{fig:nhn_overview}).
Its measurement adapter adjusts recording-dependent waveform statistics while
retaining relative intensity. A causal spatiotemporal encoder combines temporal
convolutions and parameter-efficient interactions, then modulates features using
multi-timescale activity histories. A compact projection
maps these features to non-negative candidate drives, which are causally
integrated before Henneman-inspired graded allocation forms the compact latent
neuromotor state.
The state sequence and retained intensity cues support task-specific decoding.
We provide theoretical support for these architectural choices.

We make three contributions.

\begin{itemize}
    \item We introduce NHN, a physiology-guided sEMG decoding architecture
    that infers a compact, task-aligned latent neuromotor state to support
    generalization across users and sessions.
    \item NHN combines learnable causal adaptation that preserves relative
    intensity, parameter-efficient channel interactions, and feature regulation
    based on activity histories.
    Non-negative drives prevent cancellation during integration, while
    continuous graded allocation regulates each primitive's contribution.
    \item We evaluate a common NHN architecture with task-specific prediction
    heads on emg2pose and emg2qwerty, demonstrating its applicability to both
    continuous hand-pose estimation and discrete character-sequence recognition.
\end{itemize}

%% file: sections/related_work.tex
\section{Related Work}
\label{sec:related_work}

\paragraph{Generalization in sEMG decoding.}
Population-scale training supports calibration-free interfaces
\citep{kaifosh2025generic}, while benchmarks \citep{yang2024emgbench} distinguish
intersubject transfer from adaptation and broaden demographic coverage
\citep{gowda2025diverse}. Target-domain adaptation uses voting-based pseudo-label
refinement \citep{duan2023emgsense}, memory-based labeling in spiking networks
\citep{guo2024spgesture}, and class-level alignment \citep{zhong2025ppsda}.
Disentanglement preserves shared and target-specific features
\citep{su2025msafd}, while adapter expansion and rehearsal limit forgetting
across new users \citep{chen2024crosssubjectlifelong}. Pretraining supports
few-shot personalization through self-supervised task replay \citep{yoon2025selfreplay}.
Source training learns cross-posture invariance through Fourier-phase
distillation \citep{su2025dbssdg} and reusable patterns through quantized
myoelectric codebooks \citep{huang2026aemg}. Architectures use
short-window attention for noisy gesture decoding \citep{guo2025stem} and linear
state-space models for typing \citep{ortone2026efficient}. We exclude the latter
from quantitative comparisons because evaluation protocols differ.
NHN organizes task-relevant coordination
with physiological priors in a compact decoder.

\paragraph{Transferable representations for physiological signals.}
For brain recordings, geometry-aware encodings accommodate sensor layouts
\citep{elouahidi2025reve,xiao2025brainomni}, while factorized and cross-scale
attention model spatiotemporal dependencies \citep{wang2025cbramod,zhou2025csbrain}.
Temporal-spectral tokens \citep{ma2026codebrain}, multiscale wavelets
\citep{chen2025physiowave}, and physiological frequency bands for ear-recorded
signals \citep{yoon2026beyond} structure time-frequency content.
Pretraining combines masked reconstruction with latent alignment
\citep{wang2024eegpt} or learns text-aligned tokens \citep{jiang2025neurolm}.
Optical and cardiac models exploit pulse morphology \citep{pillai2025papagei}
and electrical-to-peripheral timing \citep{zhou2026xmae}. Low-rank weight
generation personalizes models using local data \citep{wu2025efficient}.
For task-supervised sEMG, the question is how physiological priors can jointly
guide measurement processing and activity representation. NHN unifies them
in a neuromotor-inspired hierarchy.

\paragraph{Latent neuromotor states and motor primitives.}
Motor-state representations differ in scale and behavioral relevance.
Simulation benchmarks \citep{mamidanna2025muniverse} and separation algorithms
\citep{tacca2025muelim} support decomposition into individual motor-unit discharge
trains. Aligned population recordings reveal cross-animal regularities
\citep{safaie2023preserved}, while neural trajectories exhibit dynamical constraints
\citep{oby2025dynamical}. Dynamical embeddings capture cross-recording variation
\citep{vermani2025metadynamical}, input-driven models separate intrinsic dynamics
from measured-input effects \citep{vahidi2025braid}, and multi-region models distinguish
shared and region-specific subspaces \citep{sristi2026ctae}.
Behavior further constrains representations through movement-supervised neural
embeddings \citep{chen2024ner} and sparse-input dynamical models of larval zebrafish
locomotion \citep{mullen2024locomotion}. Synergy-based control uses low-rank
policies \citep{dong2022lowrank} or state-dependent synergies \citep{he2024dynsyn},
while synergy-based neural manifolds link population activity to hand kinematics
\citep{sun2025synergynet}. For sEMG, the challenge is to infer a compact,
task-relevant latent neuromotor state from mixed surface measurements.
NHN organizes its non-negative components through temporal integration and
graded allocation under task supervision.

%% file: sections/method.tex
\section{Method: Neuromotor Hierarchy Network}
\label{sec:method}
\label{sec:method-emg2qwerty}
\label{sec:summary}
\label{sec:bio_generative}

\begin{figure}[!t]
\centering
\includegraphics[width=\textwidth]{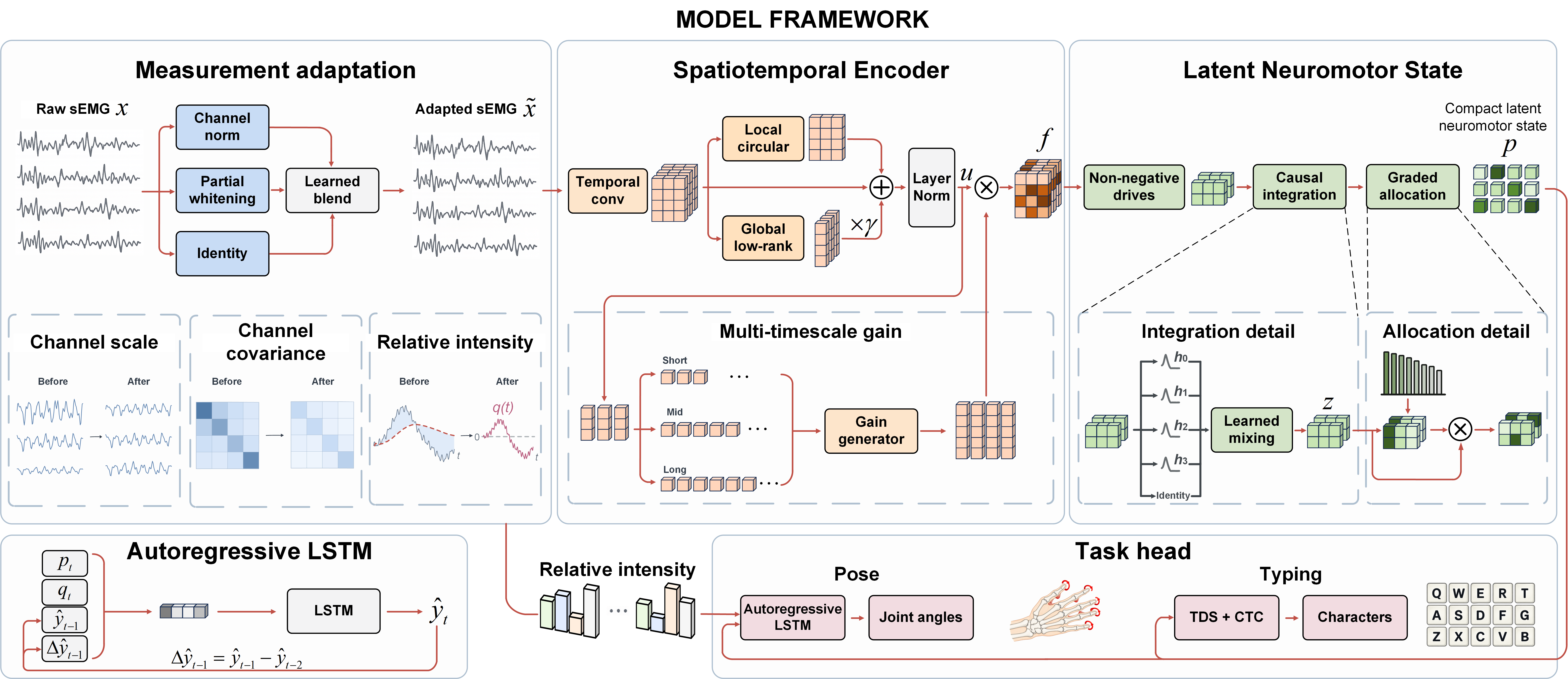}
\caption{NHN architecture.
  Task-trained NHN adapts and encodes sEMG, then integrates and allocates candidate drives
  to form a compact latent neuromotor state for pose and typing decoding.}
\label{fig:nhn_overview}
\end{figure}

Inspired by neuromotor organization, NHN learns a compact latent neuromotor
state for sEMG decoding. Measurement adaptation and spatiotemporal encoding
provide evidence of coordination, which the network organizes into graded
primitives for task-specific prediction (Figure~\ref{fig:nhn_overview}).

Let $\vx\in\mathbb{R}^{C\times T_0}$ contain $T_0$ samples from $C$ sEMG
channels, and let $\kappa$ index the task. Denote measurement adaptation by
$\mathcal{A}$, spatiotemporal encoding by $\mathcal{E}$, latent neuromotor state
construction by $\mathcal{B}$, and task-specific prediction by $\mathcal{H}_\kappa$.
For pose estimation, the data flow is
\begin{equation}
(\tilde{\vx},\vq)=\mathcal{A}(\vx),\quad
\vf=\mathcal{E}(\tilde{\vx}),\quad
\vp=\mathcal{B}(\vf),\quad
\hat{\vy}=\mathcal{H}_{\kappa}(\vp,\vq^{\downarrow}).
\label{eq:nhn_factorization}
\end{equation}
The adapted waveform $\tilde{\vx}$ and relative intensity $\vq$ retain the
input dimensions. Aligning $\vq$ to the encoder's time grid gives
$\vq^{\downarrow}\in\mathbb{R}^{C\times L_\kappa}$. The encoded sequence
$\vf\in\mathbb{R}^{D\times L_\kappa}$ has width $D$, while the latent neuromotor state
$\vp\in\mathbb{R}_{\geq0}^{N\times L_\kappa}$ contains $N$ primitives over
the same $L_\kappa$ frames.
Throughout, bold feature symbols denote sequences; a time subscript selects
one column, and two subscripts select a scalar entry. We use the rectified
linear unit (ReLU) and layer normalization (LN).

\subsection{Measurement Adaptation with Intensity Preservation}
\label{sec:asc_analysis}

Tissue filtering and electrode sampling motivate adjusting recording-dependent
scale and covariance while retaining short-term intensity cues
\citep{deluca1988crosstalk,farina2014extraction}.

For input sample $n=1,\ldots,T_0$, let $\ell_{c,n}$ be channel $c$'s log
standard deviation over a causal window ending at $n$. The intensity path
centers it against its cumulative mean. Partial whitening $\mathcal{W}$
uses the channel mean $\vmu_b$ and covariance $\mSigma_b$ of samples
preceding each block $b$,
\begin{equation}
\begin{aligned}
q_{c,n}&=\ell_{c,n}-\frac{1}{n}\sum_{i=1}^{n}\ell_{c,i},\\
\mathcal{W}(\vx)_n
&=\left[(1-\alpha)\operatorname{Diag}(\mSigma_b)
+\alpha\mSigma_b+\lambda_b\mI_C\right]^{-1/2}
(\vx_n-\vmu_b),\quad n\in b.
\end{aligned}
\label{eq:partial_zca}
\end{equation}
Here $\operatorname{Diag}$ retains the covariance diagonal and $\mI_C$ is
the identity matrix. Fixed $\alpha\in[0,1]$ interpolates between channel-wise
normalization and joint whitening without a principal-component rotation.
The regularizer $\lambda_b>0$ is computed from mean channel variance.
The first block uses cumulative diagonal normalization, with warmup confined
to the input's left context. For pose, the waveform path blends
$\mathcal{W}(\vx)$ with $\mathcal{D}(\vx)$, which normalizes using cumulative
channel-wise means and standard deviations. Both operate on the original
$\vx$, with learned residual strengths
$\eta_{\mathrm d},\eta_{\mathrm w}\in\mathbb{R}$,
\begin{equation}
\tilde{\vx}
=(1-\eta_{\mathrm w})
\left[\vx+\eta_{\mathrm d}\bigl(s_0\mathcal{D}(\vx)-\vx\bigr)\right]
+\eta_{\mathrm w}s_0\mathcal{W}(\vx).
\label{eq:measurement_blend}
\end{equation}
The once-calibrated, fixed scale $s_0>0$ puts both candidates on the input's
amplitude scale. $\vq$ retains intensity for task-dependent fusion.

\subsection{Spatiotemporal Encoding of Adapted sEMG}
\label{sec:structured_analysis}

The encoder combines temporal convolution and local-global
mixing in time-depth separable (TDS) blocks \citep{hannun2019sequence}, followed
by multi-timescale adaptive gain to produce $\vf$.

\paragraph{Temporal feature extraction.}
Within a block $r$ of width $D_r$,
$\vu^{\mathrm{in}}$, $\vu^{\mathrm{mid}}$, and $\vu^{\mathrm{out}}$ denote
its input, temporal features, and output, respectively. Convolutional input
layers map $\tilde{\vx}$ to the first block's $\vu^{\mathrm{in}}$.
Temporal convolution $\mathcal{T}_r$ produces
\begin{equation}
\vu^{\mathrm{mid}}=\operatorname{LN}\!\left(
\operatorname{crop}_r(\vu^{\mathrm{in}})+\operatorname{ReLU}\!\left(
\mathcal{T}_{r}(\vu^{\mathrm{in}})\right)\right).
\label{eq:tds_temporal}
\end{equation}
$\operatorname{crop}_r$ retains trailing input frames to match the unpadded
convolution's output length.

\paragraph{Local-global mixing of temporal features.}
Motivated by local correlations and distributed muscle coordination
\citep{farina2014extraction,davella2003combinations}, local-global mixing
combines shared local interactions with a low-rank path connecting distant
groups. The global path uses the Gaussian error linear unit (GELU).
At block frame $t$, the residual output is
\begin{equation}
\vu_t^{\mathrm{out}}=\operatorname{LN}\left(
\vu_t^{\mathrm{mid}}+\mathcal{C}_{k}^{(r)}(\vu_t^{\mathrm{mid}})
+\gamma_r\mW_{\mathrm{up}}^{(r)}
\operatorname{GELU}(\mW_{\mathrm{down}}^{(r)}\vu_t^{\mathrm{mid}})
\right).
\label{eq:circconv}
\end{equation}
The local operator $\mathcal{C}_{k}^{(r)}$ applies two width-$k$ circular
convolutions separated by ReLU, sharing weights across $G$ feature groups
arranged on a ring, each containing $D_r/G$ features. The global projections
$\mW_{\mathrm{down}}^{(r)}$ and $\mW_{\mathrm{up}}^{(r)}$ project from width
$D_r$ to $R$ and back, with a learned gate $\gamma_r$.
Projection biases are omitted in Equation~\ref{eq:circconv}.
Each block has independent parameters and passes $\vu^{\mathrm{out}}$ to the
next block's $\vu^{\mathrm{in}}$, with convolutional projection when width or
rate changes. A final projection, when needed, yields
$\vu\in\mathbb{R}^{D\times L_\kappa}$ for adaptive gain.

\paragraph{Multi-timescale contextualization.}
The mixed features $\vu$ are reweighted using causal history at multiple
timescales to capture brief and sustained activity, motivated by common
drive and muscle coordination \citep{negro2016common,davella2003combinations}.
Each of $M$ context states accumulates the same input $\vu_t$ with a
different learned timescale,
\begin{equation}
\vs^{(m)}_t=\left(1-e^{-\Delta t/\tau_m}\right)\vu_t
+e^{-\Delta t/\tau_m}\vs^{(m)}_{t-1},
\qquad m=1,\ldots,M.
\label{eq:mtag_state}
\end{equation}
On the final encoder grid $t=0,\ldots,L_\kappa-1$, the context states
$\vs_t^{(m)}\in\mathbb{R}^{D}$ use frame interval $\Delta t$, learned
$\tau_m\geq\Delta t$, and learned initial states $\vs^{(m)}_{-1}$. A learned map
$g_a:\mathbb{R}^{MD}\rightarrow\mathbb{R}^{D}$ converts the concatenated
states into a residual gain on the current features,
\begin{equation}
\vf_t=\vu_t\odot\left[1+\tanh\!\left(
g_a\!\left([\vs^{(1)}_t;\ldots;\vs^{(M)}_t]\right)\right)\right],
\label{eq:mtag_gain}
\end{equation}
where $\odot$ denotes elementwise multiplication.

\subsection{Latent Neuromotor State Construction from Encoded Features}
\label{sec:synthesis}

Through drive projection, temporal integration, and graded allocation,
$\mathcal{B}$ maps contextualized features $\vf$ into a compact latent neuromotor
state $\vp$, limiting the width passed to the prediction head.

\paragraph{Non-negative candidate drives.}
A learned pointwise affine map $g_d$, layer normalization, and Softplus
produce $N$ non-negative candidate drives per frame,
\begin{equation}
\vd_t=\operatorname{Softplus}\!\left(\operatorname{LN}(g_d(\vf_t))\right)
\in\mathbb{R}_{\geq 0}^{N}.
\label{eq:candidate_drive}
\end{equation}
Non-negativity prevents cancellation during integration. Low-dimensional coordination motivates the compact drive space
\citep{tresch1999extraction,davella2003combinations,bizzi2013neural}.

\paragraph{Temporal integration of candidate drives.}
The candidate drives $\vd$ are integrated before allocation, inspired by
synaptic input integration \citep{powers2017synaptic}. Diverse motor-unit
contraction times motivate heterogeneous temporal support
\citep{burke1973physiological,heckman2012motor}, implemented with causal
finite impulse response (FIR) paths. Combining $M_f$ filtered versions of
$\vd$ with an identity path gives
\begin{equation}
\bar{\vd}_t=\boldsymbol{\omega}_0\odot\vd_t
+\sum_{m=1}^{M_f}\boldsymbol{\omega}_m\odot
\left(\mathbf{k}_m*_{\mathrm{c}}\vd\right)_t.
\label{eq:drive_integration}
\end{equation}
Each $\mathbf{k}_m$ contains a non-negative geometric kernel per primitive,
with independently learned positive time constants. Numerical normalization
gives kernel mass at most one.
The causal operator $*_{\mathrm{c}}$ filters each primitive separately,
using left zero-padding to preserve length. Learned weights
$\boldsymbol{\omega}_m\in\mathbb{R}^{N}$ are softmax-normalized per primitive
across paths, including the identity path $m=0$. These filters integrate
candidate drives, whereas adaptive gain reweights encoder features.

\paragraph{Henneman-inspired allocation of integrated drives.}
Whereas muscle-synergy studies motivate the low-dimensional drive space,
Henneman's size principle motivates heterogeneous participation priorities
among the learned drive components
\citep{henneman1957relation,milnerbrown1973orderly}.
Behavior-dependent motor-unit control motivates activity-dependent allocation
\citep{marshall2022flexible}. NHN combines learned recruitment priorities $\boldsymbol{\beta}\in\mathbb{R}^{N}$
with the current drive at a drive-dependent temperature $\theta_t$,
\begin{equation}
\theta_t=\max\!\left\{10^{-4},\operatorname{mean}\!\left(
\operatorname{Top}_{K}(\bar{\vd}_t)\right)\right\},
\qquad
\vp_t=N\bar{\vd}_t\odot\operatorname{softmax}\!\left(
\frac{\bar{\vd}_t+\boldsymbol{\beta}}{\theta_t}\right).
\label{eq:kwta}
\end{equation}
For fixed $K\in\{1,\ldots,N\}$, $\operatorname{Top}_{K}$ returns the
$K$ largest drive entries, whose mean calibrates temperature. Softmax weights all primitives. Multiplication by $\bar{\vd}_t$ preserves continuous
magnitudes. Collecting $\vp_t$ across frames gives $\vp=\mathcal{B}(\vf)$.

\subsection{Task-specific Prediction Heads}
The prediction head $\mathcal{H}_\kappa$ maps the latent neuromotor state
sequence $\vp$ to joint-angle estimates $\hat{\vy}$ using an autoregressive long
short-term memory (LSTM) network, or to character probabilities using TDS with
connectionist temporal classification (CTC) \citep{graves2006ctc}.

%% file: sections/experiments_results.tex
\section{Experiments}
\label{sec:experiments}

We evaluate NHN on \emph{emg2pose} and \emph{emg2qwerty}, assessing
generalization across users and recording conditions alongside inference efficiency.
Both tasks use independently trained weights, encoder output width $D=64$,
and $N=32$ primitives. The backbone takes $C=16$ sEMG channels from a single
wrist at 2 kHz. It processes one wrist for pose and each wrist separately for typing.
The NHN backbone and prediction heads use causal operations.

\paragraph{Pose estimation on emg2pose.}
We use the data splits and evaluation protocol of \citet{salter2024emg2pose}. The User,
Stage, and User$\times$Stage splits hold out users, movement stages, or both.
Evaluation uses 5-s windows. Tracking receives a ground-truth initial pose,
whereas Regression does not. An autoregressive LSTM
head predicts joint angles at the 50-Hz rollout rate used by vemg2pose in \citet{salter2024emg2pose}.
We train separate Regression and Tracking models with joint supervision,
adopting the task-weighting scheme of \citet{hadidi2026reevaluating}. Metrics are angular error (AE) in degrees
and landmark distance (LD) in millimeters.

\paragraph{Typing recognition on emg2qwerty.}
We use emg2qwerty \citep{sivakumar2024emg2qwerty} with SplashNet's data splits
and evaluation settings \citep{hadidi2025splashnet}.
We pretrain NHN on pooled data from training users. On held-out users,
we evaluate both the pretrained model without adaptation (zero-shot)
and models obtained through subject-specific fine-tuning.
Other-domain validation (ODV) uses separate validation users, whereas
test-domain validation (TDV) and test-domain test (TDT) use validation and
test sessions of held-out test users. Fine-tuned comparisons use TDV and TDT.
We use a 100-Hz latent neuromotor state sequence to balance temporal resolution
and computational cost, consistent with the feature rate used by
\citet{mehlman2025scaling}. NHN shares backbone weights between wrists and
concatenates their state sequences for a TDS prediction head. Training uses CTC loss.
We report character error rate (CER)
under greedy decoding and beam search with a 6-gram language model (LM)
trained on WikiText-103.

\paragraph{Training and statistics.}
For pose training and typing pretraining, we train NHN from scratch using adaptive moment
estimation with decoupled weight decay (AdamW), cosine
learning-rate schedules, and five random seeds. We report the mean
$\pm$ standard deviation (SD) across participants after averaging each
participant's scores over seeds. Participant-level baseline statistics are
reproduced as published.

\begin{table}[t]
\centering
\setlength{\tabcolsep}{1.1pt}
\newcommand{\resultpm}[2]{#1$\pm$#2}
\caption{emg2pose Regression and Tracking. Split columns report user mean
$\pm$ SD. Session columns weight sessions within each split, then splits
equally.
Bold marks column minima within each setting, excluding
Static for efficiency. Dashes denote unreported values.
Published split results: \textsuperscript{a}\citet{hadidi2026reevaluating},
\textsuperscript{b,c,d}\citet{salter2024emg2pose}.}
\label{tab:main_emg2pose}
\makeatletter
{\edef\@currentlabel{\thetable(a)}\label{tab:main_emg2pose_reg}}
{\edef\@currentlabel{\thetable(b)}\label{tab:main_emg2pose_track}}
\makeatother
\fontsize{8}{9.5}\selectfont
\renewcommand{\arraystretch}{1.15}
\begin{tabular*}{\linewidth}{@{\extracolsep{\fill}}l@{\hspace{2.7pt}}rr@{\hspace{2.7pt}}rr@{\hspace{2.7pt}}rr@{\hspace{2.7pt}}cc@{\hspace{2.7pt}}cc@{}}
\toprule
 & \multicolumn{2}{c}{User} & \multicolumn{2}{c}{Stage}
 & \multicolumn{2}{c}{User$\times$Stage} & \multicolumn{2}{c}{Session}
 & \multicolumn{2}{c}{Efficiency} \\
\cmidrule(lr){2-3}\cmidrule(lr){4-5}\cmidrule(lr){6-7}\cmidrule(lr){8-9}\cmidrule(lr){10-11}
Model & \makecell{AE$\downarrow$\\($^\circ$)} & \makecell{LD$\downarrow$\\(mm)}
 & \makecell{AE$\downarrow$\\($^\circ$)} & \makecell{LD$\downarrow$\\(mm)}
 & \makecell{AE$\downarrow$\\($^\circ$)} & \makecell{LD$\downarrow$\\(mm)}
 & \makecell{AE$\downarrow$\\($^\circ$)} & \makecell{LD$\downarrow$\\(mm)}
 & \makecell{Params.\\($10^6$)$\downarrow$} & \makecell{GFLOPs\\per $5$ s$\downarrow$} \\
\midrule
\multicolumn{11}{@{}l}{\textbf{(a) Regression}} \\
\addlinespace[2pt]
Static\textsuperscript{a} & \resultpm{16.86}{1.80} & \resultpm{25.17}{2.84} & \resultpm{19.18}{1.71} & \resultpm{28.57}{2.46} & \resultpm{18.87}{2.01} & \resultpm{28.77}{2.78} & 17.87 & 26.81 & 0 & 0 \\
\makecell[l]{Sensing\\Dynamics\textsuperscript{b}} & \resultpm{15.50}{1.40} & \resultpm{21.80}{2.10} & \resultpm{18.80}{1.60} & \resultpm{26.60}{2.00} & \resultpm{18.70}{1.60} & \resultpm{27.20}{2.00} & -- & -- & \multicolumn{1}{c}{--} & \multicolumn{1}{c}{--} \\
NeuroPose\textsuperscript{c} & \resultpm{13.20}{1.10} & \resultpm{17.50}{1.30} & \resultpm{17.20}{1.70} & \resultpm{24.00}{2.10} & \resultpm{17.50}{1.50} & \resultpm{24.90}{1.70} & 15.52 & 21.61 & 6.35 & \textbf{1.12} \\
\addlinespace[1.5pt]
emg2pose\textsuperscript{d} & \resultpm{12.57}{1.30} & \resultpm{16.28}{1.81} & \resultpm{15.17}{1.59} & \resultpm{20.53}{2.13} & \resultpm{15.57}{1.30} & \resultpm{21.49}{1.71} & 14.01 & 18.94 & \textbf{2.95} & 2.21 \\
vemg2pose\textsuperscript{d} & \resultpm{12.24}{1.34} & \resultpm{15.82}{1.90} & \resultpm{15.22}{1.58} & \resultpm{20.38}{2.07} & \resultpm{15.63}{1.37} & \resultpm{21.30}{1.87} & 13.91 & 18.69 & 5.98 & 3.72 \\
\addlinespace[1.5pt]
Pos-ST\textsuperscript{a} & \resultpm{12.19}{1.27} & \resultpm{15.79}{1.80} & \resultpm{14.96}{1.63} & \resultpm{20.23}{2.15} & \resultpm{15.47}{1.30} & \resultpm{21.29}{1.82} & -- & -- & 5.97 & 3.72 \\
Vel-ST\textsuperscript{a} & \resultpm{12.17}{1.31} & \resultpm{15.58}{1.85} & \resultpm{14.62}{1.65} & \resultpm{19.62}{2.21} & \resultpm{15.35}{1.32} & \resultpm{20.92}{1.87} & -- & -- & 5.98 & 3.72 \\
Pos-MT\textsuperscript{a} & \resultpm{11.54}{1.19} & \resultpm{14.74}{1.64} & \resultpm{14.02}{1.62} & \resultpm{18.61}{2.16} & \resultpm{14.58}{1.29} & \resultpm{19.78}{1.83} & -- & -- & 5.97 & 3.72 \\
Vel-MT\textsuperscript{a} & \resultpm{11.64}{1.25} & \resultpm{14.74}{1.74} & \resultpm{13.85}{1.62} & \resultpm{18.33}{2.16} & \resultpm{14.63}{1.34} & \resultpm{19.77}{1.95} & -- & -- & 5.98 & 3.72 \\
\addlinespace[1.5pt]
\textbf{NHN (Ours)} & \resultpm{\textbf{11.38}}{1.13} & \resultpm{\textbf{14.42}}{1.53} & \resultpm{\textbf{13.78}}{1.57} & \resultpm{\textbf{18.21}}{2.06} & \resultpm{\textbf{14.22}}{1.26} & \resultpm{\textbf{19.26}}{1.88} & \textbf{12.73} & \textbf{16.88} & 3.08 & 2.40 \\
\midrule
\multicolumn{11}{@{}l}{\textbf{(b) Tracking}} \\
\addlinespace[2pt]
Static\textsuperscript{a} & \resultpm{10.48}{1.81} & \resultpm{15.71}{2.98} & \resultpm{14.86}{1.91} & \resultpm{21.81}{2.58} & \resultpm{14.29}{1.86} & \resultpm{21.48}{2.33} & 13.16 & 19.99 & 0 & 0 \\
\addlinespace[1.5pt]
emg2pose\textsuperscript{d} & \resultpm{11.33}{1.10} & \resultpm{14.75}{1.53} & \resultpm{15.39}{1.50} & \resultpm{20.88}{1.92} & \resultpm{15.05}{1.22} & \resultpm{20.69}{1.39} & 13.44 & 18.23 & \textbf{2.95} & \textbf{2.21} \\
vemg2pose\textsuperscript{d} & \resultpm{7.70}{0.98} & \resultpm{10.22}{1.54} & \resultpm{11.20}{1.43} & \resultpm{15.18}{1.81} & \resultpm{11.03}{1.01} & \resultpm{15.35}{1.46} & 9.84 & 13.56 & 5.97 & 3.72 \\
\addlinespace[1.5pt]
Pos-ST\textsuperscript{a} & \resultpm{7.35}{0.92} & \resultpm{9.67}{1.48} & \resultpm{10.21}{1.32} & \resultpm{13.73}{1.69} & \resultpm{10.25}{0.93} & \resultpm{14.18}{1.35} & -- & -- & 5.97 & 3.72 \\
Vel-ST\textsuperscript{a} & \resultpm{7.64}{0.94} & \resultpm{10.02}{1.52} & \resultpm{10.76}{1.34} & \resultpm{14.43}{1.71} & \resultpm{10.75}{0.93} & \resultpm{14.82}{1.40} & -- & -- & 5.97 & 3.72 \\
Pos-MT\textsuperscript{a} & \resultpm{7.36}{0.90} & \resultpm{9.68}{1.46} & \resultpm{10.28}{1.33} & \resultpm{13.81}{1.70} & \resultpm{10.29}{0.97} & \resultpm{14.22}{1.37} & -- & -- & 5.97 & 3.72 \\
Vel-MT\textsuperscript{a} & \resultpm{7.64}{0.95} & \resultpm{10.00}{1.53} & \resultpm{10.77}{1.36} & \resultpm{14.44}{1.75} & \resultpm{10.75}{0.97} & \resultpm{14.81}{1.41} & -- & -- & 5.97 & 3.72 \\
\addlinespace[1.5pt]
\textbf{NHN (Ours)} & \resultpm{\textbf{7.30}}{0.88} & \resultpm{\textbf{9.55}}{1.41} & \resultpm{\textbf{10.12}}{1.31} & \resultpm{\textbf{13.59}}{1.69} & \resultpm{\textbf{10.16}}{0.99} & \resultpm{\textbf{14.03}}{1.33} & \textbf{9.08} & \textbf{12.39} & 3.08 & 2.40 \\
\bottomrule
\end{tabular*}
\end{table}

\begin{table}[t]
\centering
\setlength{\tabcolsep}{0.7pt}
\newcommand{\resultpm}[2]{#1$\pm$#2}
\newcommand{\resultplain}[1]{\multicolumn{1}{c}{#1}}
\caption{emg2qwerty CER (\%). NHN reports mean $\pm$ SD across
participants. Bold marks column minima within
each setting. Dashes denote unreported results. ODV is not evaluated
after fine-tuning.
\textsuperscript{a}\citet{sivakumar2024emg2qwerty} (ODV and efficiency from \textsuperscript{b}).
\textsuperscript{b}\citet{hadidi2025splashnet}.
\textsuperscript{c}\citet{mehlman2025scaling}.}
\label{tab:main_emg2qwerty}
\makeatletter
{\edef\@currentlabel{\thetable(a)}\label{tab:main_emg2qwerty_zs}}
{\edef\@currentlabel{\thetable(b)}\label{tab:main_emg2qwerty_ft}}
\makeatother
\fontsize{8}{9.5}\selectfont
\renewcommand{\arraystretch}{1.15}
\begin{tabular*}{\linewidth}{@{\extracolsep{\fill}}l@{\hspace{3pt}}rr@{\hspace{3pt}}rr@{\hspace{3pt}}rr@{\hspace{3pt}}cc@{}}
\toprule
 & \multicolumn{2}{c}{ODV} & \multicolumn{2}{c}{TDV}
 & \multicolumn{2}{c}{TDT} & \multicolumn{2}{c}{Efficiency} \\
\cmidrule(lr){2-3}\cmidrule(lr){4-5}\cmidrule(lr){6-7}\cmidrule(lr){8-9}
Model & \makecell{Greedy$\downarrow$\\(\%)} & \makecell{Beam$\downarrow$\\(\%)}
 & \makecell{Greedy$\downarrow$\\(\%)} & \makecell{Beam$\downarrow$\\(\%)}
 & \makecell{Greedy$\downarrow$\\(\%)} & \makecell{Beam$\downarrow$\\(\%)}
 & \makecell{Params.$\downarrow$\\($10^6$)} & \makecell{GFLOPs$\downarrow$\\per $30$ s} \\
\midrule
\multicolumn{9}{@{}l}{\textbf{(a) Zero-shot}} \\
\addlinespace[2pt]
TDS ConvNet\textsuperscript{a} & \resultplain{72.44} & \resultplain{72.07} & \resultplain{55.57} & \resultpm{52.10}{5.54} & \resultplain{55.38} & \resultpm{51.78}{4.61} & 5.29 & 61.61 \\
\addlinespace[1.5pt]
\makecell[l]{SplashNet\\(Split-only)\textsuperscript{b}} & \resultplain{61.74} & \resultplain{58.64} & \resultplain{45.73} & \resultpm{37.28}{6.91} & \resultplain{45.69} & \resultpm{37.37}{7.34} & 2.68 & 36.84 \\
\makecell[l]{SplashNet-mini\\(Shared)\textsuperscript{b}} & \resultplain{61.07} & \resultplain{58.20} & \resultplain{45.33} & \resultpm{36.46}{7.09} & \resultplain{45.26} & \resultpm{36.41}{7.30} & 1.38 & 36.84 \\
SplashNet-Upscale\textsuperscript{b} & \resultplain{60.16} & \resultplain{56.95} & \resultplain{44.79} & \resultpm{35.49}{7.56} & \resultplain{44.78} & \resultpm{35.67}{6.79} & 2.58 & 71.38 \\
\addlinespace[1.5pt]
Transformer-tiny\textsuperscript{c} & \multicolumn{1}{c}{--} & \multicolumn{1}{c}{--} & \multicolumn{1}{c}{--} & \multicolumn{1}{c}{--} & \resultplain{\textbf{31.90}} & \multicolumn{1}{c}{--} & 2.20 & $\sim$9720 \\
Transformer-small\textsuperscript{c} & \multicolumn{1}{c}{--} & \multicolumn{1}{c}{--} & \multicolumn{1}{c}{--} & \multicolumn{1}{c}{--} & \resultplain{32.70} & \multicolumn{1}{c}{--} & 5.40 & $\sim$18720 \\
\addlinespace[1.5pt]
\textbf{NHN (Ours)} & \resultpm{\textbf{56.93}}{13.82} & \resultpm{\textbf{51.73}}{15.32} & \resultpm{\textbf{39.03}}{7.08} & \resultpm{\textbf{29.29}}{8.99} & \resultpm{38.62}{7.01} & \resultpm{\textbf{28.75}}{8.29} & \textbf{0.88} & \textbf{17.85} \\
\midrule
\multicolumn{9}{@{}l}{\textbf{(b) Subject-specific fine-tuning}} \\
\addlinespace[2pt]
\multicolumn{3}{@{}l}{TDS ConvNet\textsuperscript{a}} & \resultplain{11.39} & \resultpm{8.31}{3.19} & \resultplain{11.28} & \resultpm{6.95}{3.61} & 5.29 & 61.61 \\
\addlinespace[1.5pt]
\multicolumn{3}{@{}l}{SplashNet (Split-only)\textsuperscript{b}} & \resultplain{12.67} & \resultpm{6.04}{2.81} & \resultplain{12.89} & \resultpm{6.10}{3.32} & 2.68 & 36.84 \\
\multicolumn{3}{@{}l}{SplashNet-mini (Shared)\textsuperscript{b}} & \resultplain{13.01} & \resultpm{6.13}{2.96} & \resultplain{13.22} & \resultpm{5.87}{3.04} & 1.38 & 36.84 \\
\multicolumn{3}{@{}l}{SplashNet-mini (Unshared)\textsuperscript{b}} & \resultplain{12.68} & \resultpm{5.85}{2.83} & \resultplain{13.07} & \resultpm{5.96}{3.28} & 2.68 & 36.84 \\
\multicolumn{3}{@{}l}{SplashNet-Upscale (Shared)\textsuperscript{b}} & \resultplain{12.10} & \resultpm{5.46}{2.60} & \resultplain{12.39} & \resultpm{5.51}{2.81} & 2.58 & 71.38 \\
\multicolumn{3}{@{}l}{SplashNet-Upscale (Unshared)\textsuperscript{b}} & \resultplain{11.80} & \resultpm{5.57}{2.65} & \resultplain{12.13} & \resultpm{5.67}{2.97} & 5.06 & 71.38 \\
\addlinespace[1.5pt]
\multicolumn{3}{@{}l}{Transformer-tiny\textsuperscript{c}} & \multicolumn{1}{c}{--} & \multicolumn{1}{c}{--} & \resultplain{8.60} & \multicolumn{1}{c}{--} & 2.20 & $\sim$9720 \\
\multicolumn{3}{@{}l}{Transformer-small\textsuperscript{c}} & \multicolumn{1}{c}{--} & \multicolumn{1}{c}{--} & \resultplain{\textbf{7.10}} & \multicolumn{1}{c}{--} & 5.40 & $\sim$18720 \\
\addlinespace[1.5pt]
\multicolumn{3}{@{}l}{\textbf{NHN (Ours)}} & \resultpm{\textbf{9.12}}{3.76} & \resultpm{\textbf{3.95}}{1.62} & \resultpm{9.41}{3.51} & \resultpm{\textbf{3.83}}{1.51} & \textbf{0.88} & \textbf{17.85} \\
\bottomrule
\end{tabular*}
\end{table}

\paragraph{Baselines and efficiency.}
The emg2pose baseline shares the dataset's name and directly predicts
joint angles \citep{salter2024emg2pose}.
Other pose baselines include vemg2pose, NeuroPose
\citep{liu2021neuropose}, Sensing Dynamics \citep{simpetru2022sensing}, the position (Pos) and velocity (Vel), single-task (ST) and multi-task (MT)
variants of \citet{hadidi2026reevaluating},
and Static. These pose comparisons use published results under matching
evaluation protocols. We additionally
evaluate available pretrained checkpoints at the session level.
Typing comparisons use the TDS convolutional network (TDS ConvNet)
\citep{sivakumar2024emg2qwerty},
SplashNet \citep{hadidi2025splashnet}, and distilled Transformers
\citep{mehlman2025scaling}.
We report model parameters (Params.) and floating-point operations (FLOPs), expressed
in billions (GFLOPs), for 5-s pose spans and 30-s typing inputs.
These counts cover neural-network inference, including prediction heads,
and exclude language-model and text-decoding costs.

\section{Results and Discussion}
\label{sec:results}

\subsection{Main results}
\label{sec:main_results}

On \emph{emg2pose}, NHN's advantage over vemg2pose is more pronounced
when generalization includes unseen movement stages
(Table~\ref{tab:main_emg2pose}). In both Regression and Tracking,
absolute reductions in user-level AE and LD are larger on Stage and
User$\times$Stage than on User. For example, Regression AE decreases by
$1.44^\circ$ on Stage compared with $0.86^\circ$ on User.
Session-equal aggregation within each split preserves the advantage: AE decreases by
$8.48\%$ in Regression and $7.73\%$ in Tracking, with corresponding LD
improvements. Against the stronger variants of
\citet{hadidi2026reevaluating}, which retain the vemg2pose core architecture
with revised decoding and training, NHN achieves lower user-level AE and LD with $48.42\%$ to $48.51\%$
fewer parameters and $35.55\%$ to $35.64\%$ less computation. Hadidi et al.'s
best-performing variants differ between tasks (Vel-MT for Regression and
Pos-ST for Tracking), whereas NHN uses a common architecture and
joint-supervision formulation.

On \emph{emg2qwerty}, NHN's gains are present before language-model-assisted
decoding and persist after individual adaptation
(Table~\ref{tab:main_emg2qwerty}). On TDT, greedy CER decreases by
$6.16$ percentage points relative to \mbox{SplashNet-Upscale}
\citep{hadidi2025splashnet} in zero-shot evaluation and by $2.98$ points
relative to its Shared variant after fine-tuning. Relative beam-search CER
reductions against these baselines are $19.40\%$ and $30.42\%$, respectively.
NHN outperforms TDS ConvNet and all SplashNet variants across the evaluated
splits, settings, and decoding methods. These gains accompany $65.86\%$ fewer
parameters and $74.99\%$ less neural-network computation than SplashNet-Upscale.
The Transformer baselines achieve lower TDT greedy CER, but with larger
parameter counts and greater computation.

\subsection{Learned organization of the latent neuromotor state}
\label{sec:mech_interp}

\begin{figure}[!t]
\centering
\includegraphics[width=0.80\textwidth]{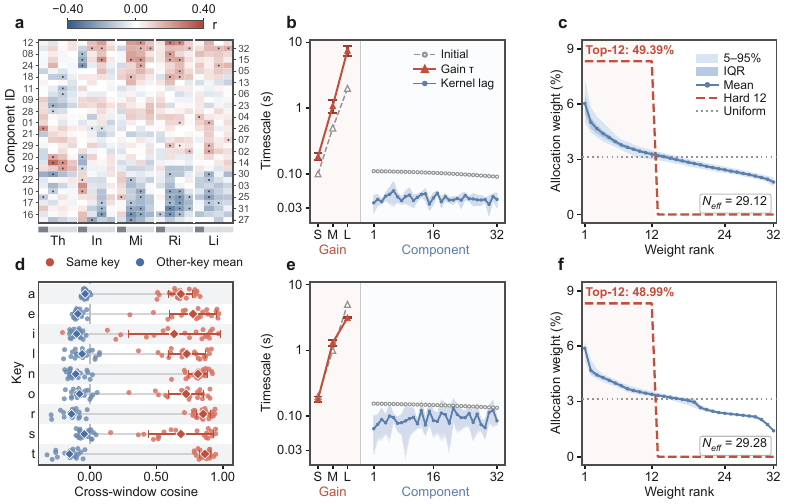}
\caption{Learned organization of the compact latent neuromotor state.
  Pose Regression: (a) participant-mean Pearson correlations between relative
  primitive activations and joint angles.
  Dots mark $\geq80\%$ participant agreement with the mean sign.
  Dark/light strips indicate abduction-adduction/flexion-extension.
  Th/In/Mi/Ri/Li denote thumb/index/middle/ring/little fingers.
  (b) Initial and post-training adaptive-gain timescales
  ($\tau_m$, Equation~\ref{eq:mtag_state}) and mean integration lags,
  kernel-weighted mean ages of candidate drives
  (Equation~\ref{eq:drive_integration}), with five-run mean$\pm$SD.
  S, M, and L denote short, intermediate, and long timescales.
  (c) Mean within-frame ranked allocation weights, with dark interquartile range (IQR)
  and light 5-95\% frame bands.
  Hard-12/Uniform assign equal weights to 12/32 primitives.
  Typing: (d) within-participant cross-window key-profile cosine similarity.
  Circles denote participant-model observations, diamonds means,
  bars participant SD.
  (e) Initial and post-training adaptive-gain timescales and mean integration
  lags (five-run mean$\pm$SD).
  (f) Ranked allocation weights with bands and reference
  distributions as in (c).}
\label{fig:fig2}
\end{figure}

Figure~\ref{fig:fig2} shows how NHN's physiology-inspired hierarchy
learns different patterns in the compact latent neuromotor state for pose
estimation and typing. Its primitives (Section~\ref{sec:synthesis}) are the
individual state dimensions, labeled components in the figure. Relative
activation is each primitive's share of the summed activation across primitives.

On emg2pose, relative primitive activations have overlapping associations
with joint angles across fingers in the sampled validation data
(Figure~\ref{fig:fig2}a). These associations suggest that hand posture is
represented through the combined activity of multiple primitives,
consistent with NHN's coordination-inspired design.
Training increases all three adaptive-gain timescales but decreases mean
integration lags, the kernel-weighted average age of candidate drives
(Figure~\ref{fig:fig2}b). For continuous pose estimation,
this suggests using longer context to reweight encoder features through
adaptive gain, while emphasizing recent candidate drives when constructing
the state.
Allocation weights set the relative emphasis on integrated candidate drives
(Equation~\ref{eq:kwta}). They remain spread across primitives, with the 12
largest accounting for about half the total allocation weight in the
illustrated clip (Figure~\ref{fig:fig2}c).

On emg2qwerty, primitive-activation patterns across the sampled training
windows show higher mean similarity for the same key than for different
keys within participants, across all nine
analyzed keys (Figure~\ref{fig:fig2}d). These recurring patterns suggest
that repeated keystrokes reuse similar primitive combinations.
Training decreases the short (S) and long (L) adaptive-gain timescales but
increases the intermediate (M) one, while mean integration lags decrease
(Figure~\ref{fig:fig2}e). For discrete typing events, this suggests using
intermediate context for adaptive gain while emphasizing recent candidate
drives during state construction, rather than extending history at every
gain scale.
Allocation is also spread across primitives with unequal
weights in the illustrated typing clip (Figure~\ref{fig:fig2}f).
Together, these patterns suggest that NHN learns to weigh recent and earlier
activity differently for adaptive gain and primitive-wise drive
integration, with task-specific preferences within the same architecture.

On the sampled validation data, uniform allocation, with equal allocation weights
across primitives, increases pose AE by
$3.94^\circ$ and typing greedy CER by 42.12 percentage points compared with
learned allocation. These increases occur
at unchanged total state activation, indicating that the prediction heads
rely on its distribution across primitives, not just its overall strength.

\section{Limitations}
\label{sec:limitations}

NHN's learned components are task representations rather than genuine
physiological sources. Generalization is evaluated on wrist sEMG.
Physical-device deployment and transfer to other sensor layouts remain untested.
At longer fixed tracking horizons, NHN's errors are higher, although
they remain below vemg2pose's and the gap between them is larger.

\section{Conclusion}
\label{sec:conclusion}

NHN uses a physiology-inspired hierarchy to infer a compact latent neuromotor
state from wrist sEMG for hand-pose estimation and touch-typing recognition.
The learned organization
takes different forms: distributed joint-angle associations in pose and
recurring key-specific combinations in typing, accompanied by distinct
temporal preferences. Allocation interventions show that decoding uses
relative activation patterns beyond total intensity. Theoretical analysis
characterizes the underlying computational mechanisms.
Together with the generalization and efficiency results, these
findings support an architectural approach in which physiological organization
guides representation structure while task supervision shapes its content.